\documentclass[11pt]{article}

\pdfoutput=1

\usepackage{lineno,hyperref}
\modulolinenumbers[5]

\usepackage{amsmath}
\usepackage{amssymb}
\usepackage{amsthm}
\usepackage{tikz}
\usepackage{pgfplots}
\usepackage{slashbox}
\usepackage{caption}
\usepackage{subcaption}
\usepackage{xcolor}
\usepackage{balance}
\allowdisplaybreaks  
\usepackage{cite}
\usepackage{algorithm}
\usepackage[noend]{algcompatible}
\usepackage{mathtools}
\usepackage{subcaption}
\usepackage{comment}
\usepackage{tabularx}
\usepackage{bbm}
\usepackage{thmtools}

\allowdisplaybreaks

\renewcommand{\thmcontinues}[1]{cont.}

\usepackage{array}
\usepackage{authblk}
\usepackage{setspace}
\usepackage{fullpage,etoolbox}

\newtheorem{theorem}{Theorem}
\newtheorem{lemma}{Lemma}
\newtheorem{assumption}{Assumption}

\newtheorem{remark}{Remark}
\newtheorem{problem}{Problem} 

\newcommand{\vertiii}[1]{{\vert\kern-0.25ex\vert\kern-0.25ex\vert #1\vert\kern-0.25ex\vert\kern-0.25ex\vert}}

\newtoggle{draft}
\togglefalse{draft}

\title{Safe Planning in Dynamic Environments\\ using Conformal Prediction }
\author[1]{Lars Lindemann\thanks{Lars Lindemann and Matthew Cleaveland contributed equally.}}
\author[2]{Matthew Cleaveland$^*$}
\author[2]{Gihyun Shim}
\author[2]{George J. Pappas}
\affil[1]{Thomas Lord Department of Computer Science, University of Southern California}
\affil[2]{Department of Electrical and Systems Engineering, University of Pennsylvania}

\begin{document}

\maketitle

\begin{abstract}
We propose a framework for planning in unknown dynamic environments with probabilistic safety guarantees using conformal prediction. Particularly, we design a model predictive controller (MPC) that uses i) trajectory predictions of the dynamic environment, and ii) prediction regions quantifying the uncertainty of the predictions. To obtain prediction regions, we use conformal prediction, a statistical tool for uncertainty quantification, that requires availability of offline trajectory data -- a reasonable assumption in many applications such as autonomous driving. The prediction regions are valid, i.e., they hold with a user-defined probability, so that the MPC is provably safe.  We illustrate the results in the self-driving car simulator CARLA at a pedestrian-filled intersection. The strength of our approach is compatibility with state of the art trajectory predictors, e.g., RNNs and LSTMs, while making no assumptions on the underlying trajectory-generating distribution. To the best of our knowledge, these are the first results that provide valid safety guarantees in such a setting.
\end{abstract}


\section{Introduction}
\label{sec:introduction}

\begin{figure}
		\centering
		\includegraphics[scale=0.2]{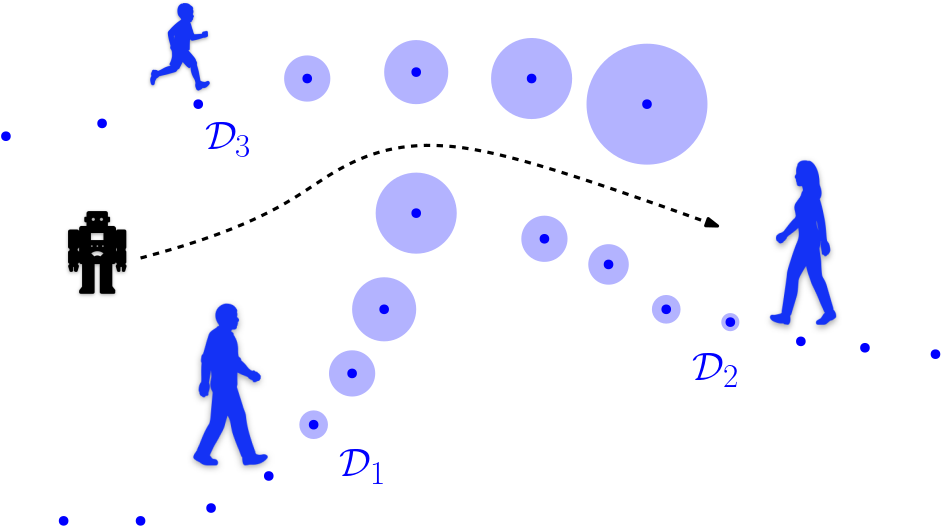}
		\caption{We predict agent trajectories using state of the art prediction algorithms, such as LSTMs, and calculate valid prediction regions (blue circles) using conformal prediction.}
		\label{fig:intro_figure}
		\vspace{-0.6cm}
\end{figure}

Mobile robots and autonomous systems  operate in dynamic and shared environments. Consider for instance a self-driving car navigating through urban traffic, or a service robot planning a path while avoiding other agents such as pedestrians, see Fig.~\ref{fig:intro_figure}. These  applications are safety-critical and challenging as the agents' intentions and policies are unknown so that their a-priori unknown trajectories need to be estimated and integrated into the planning algorithm. We propose an uncertainty-informed planning algorithm that enjoys formal safety guarantees by using conformal prediction.

The problem of path planning in dynamic environments has found broad attention \cite{mavrogiannis2021core}. A large body of work focused on multi-agent navigation without incorporating predicted agent trajectories, e.g., the dynamic window approach \cite{fox1997dynamic,mitsch2013provably} or navigation functions \cite{rimon1992exact,dimarogonas2006feedback,tanner2003nonholonomic}. However, predicted trajectories provide additional information and can significantly increase the quality of the robot's path in terms of safety and performance. Existing works that use trajectory predictions can be broadly classified into two categories: non-interactive and interactive. Non-interactive approaches predict agent trajectories and then integrate predictions into the planning algorithm \cite{trautman2010unfreezing,du2011robot}. Interactive approaches simultaneously predict agent trajectories and design the path to take the coupling effect between a control action and the trajectories of other agents into account \cite{kretzschmar2016socially,everett2021collision}. While interactive approaches attempt to model interactions between actions and agents, this is generally a difficult task and existing works fail to provide quantifiable safety guarantees.

In this paper, we focus on designing non-interactive planning algorithms with valid safety guarantees. Particularly, we use statistical tools from the conformal prediction literature \cite{vovk2005algorithmic,shafer2008tutorial} to obtain valid prediction regions that quantify the uncertainty of trajectory predictions. We then formulate a model predictive controller (MPC) that incorporates trajectory  predictions and valid prediction regions. While our framework is compatible with any trajectory prediction algorithm, we focus in the experiments on long short term memory (LSTM) networks \cite{rasouli2019pie,hochreiter1997long,rhinehart2019precog,alahi2016social,kothari2021human,salzmann2020trajectron} which are special recurrent neural networks (RNN) that can capture nonlinear and long-term trends \cite{lipton2015critical,rudenko2020human}.   Our contributions are:
\begin{itemize}
    \item We propose a planning algorithm that incorporates trajectory predictions and valid prediction regions which are obtained using conformal prediction. The elegance in using conformal prediction is that prediction regions are easy to obtain and tight. Our algorithm is computationally tractable and, under reasonable assumptions, based on a convex optimization problem.
    \item We provide valid safety guarantees which guarantee that the system is safe with a user-defined probability. Larger user-defined probabilities naturally result in more conservative plans. The strength of our approach is compatibility with state of the art trajectory predictors, e.g., RNNs and LSTMs, while making no assumptions on the underlying trajectory-generating distribution.
    \item We provide numerical experiments of a mobile robot and a self-driving car using the TrajNet++ toolbox \cite{kothari2021human} and the autonomous driving simulator CARLA~\cite{Dosovitskiy17}.
\end{itemize}

\subsection{Related Work}

The works in  \cite{phillips2011sipp,renganathan2022risk,aoude2013probabilistically,majd2021safe,kantaros2022} present non-interactive sampling-based motion planners, while \cite{du2011robot,wei2022moving,wang2022group,thomas2021probabilistic} propose non-interactive receding horizon planning algorithms that minimize the risk of collision. A challenge in non-interactive methods is the robot freezing problem in which robots may come to a deadlock due to too large prediction uncertainty \cite{trautman2010unfreezing}, e.g., for long time horizons. While this problem can be alleviated by receding horizon planning strategies, another direction to address this problem is to model social interaction as in interactive approaches where typically \textcolor{black}{an interaction model is learned or constructed \cite{kretzschmar2016socially,trautman2013robot,kuderer2012feature,mavrogiannis2020implicit,hu2022active,muvvala2022let,zhu2021learning,li2020graph,wang2020mobile}}. Reinforcement learning approaches that take social interaction into account were presented in \cite{chen2017socially,everett2021collision}. 

A particular challenge lies in selecting a good predictive model. Recent works have used intent-driven models for human agents where model uncertainty was estimated using Bayesian inference and then used for planning \cite{fisac2018probabilistically,fridovich2020confidence,bansal2020hamilton}. Other works considered Gaussian processes as a predictive model \cite{choi2017real,omainska2021gaussian,fulgenzi2008probabilistic}. To the best of our knowledge, none of the aforementioned works provide valid safety guarantees unless strong assumptions are placed on the prediction algorithm and the agent model or its distribution, e.g., being Gaussian.

We focus instead on the predictive strength of neural networks. Particularly, RNNs and LSTMs have shown to be applicable to time-series forecasting \cite{lipton2015critical,rudenko2020human}. They were successfully applied in domains such as speech/handwriting recognition and image classification \cite{graves2013speech,graves2008offline,du2021cert,arjovsky2016unitary}, but also in trajectory prediction \cite{rasouli2019pie,hochreiter1997long,rhinehart2019precog,alahi2016social,kothari2021human,salzmann2020trajectron}. We will specifically use the social LSTM presented in \cite{alahi2016social} that can jointly predict agent trajectories by taking social interaction into account. 

Neural network predictors, however, provide no information about the uncertainty of a prediction so that wrong predictions can lead to unsafe decisions. Therefore,  monitors were constructed in \cite{farid2022task,luo2021sample} to detect prediction failures -- particularly \cite{luo2021sample}  used conformal prediction to obtain guarantees on the predictor's false negative rate. Conformal prediction was also used to estimate reachable sets via neural network predictors  \cite{dietterich2022conformal,bortolussi2019neural,fan2020statistical}.  Conceptually closest to our work is \cite{chen2021reactive} where a valid predictor is constructed using conformal prediction, and then utilized to design a model predictive controller. However, no safety guarantees for the planner can be provided as the predictor uses a finite collection of training trajectories to represent all possible trajectories, implicitly requiring training and test trajectories to be similar. Our approach directly predicts trajectories of the dynamic environment (e.g., using RNNs or LSTMs) along with valid prediction regions so that we can provide end-to-end safety guarantees for our planner.


\section{Problem Formulation}
\label{sec:rnn}

We first define the safe planning problem in dynamic environments that we consider, and then briefly discuss methods for trajectory prediction of dynamic agents.

\subsection{Safe Planning in Dynamic Environments}

Consider the discrete-time dynamical system
\begin{align}
\label{eq:system}
    x_{t+1}=f(x_t,u_t), \;\;\; x_0:=\zeta
\end{align}
where $x_t\in\mathcal{X}\subseteq\mathbb{R}^n$ and $u_t\in \mathcal{U}\subseteq \mathbb{R}^m$ denote the state and the control input at time $t\in\mathbb{N}\cup \{0\}$, respectively. The sets $\mathcal{U}$ and $\mathcal{X}$ denote the set of permissible control inputs and the workspace of the system, respectively. The measurable function $f:\mathbb{R}^n\times\mathbb{R}^m\to \mathbb{R}^n$ describes the system dynamics and $\zeta\in\mathbb{R}^n$ is the initial condition of the system.

The system operates in an environment with $N$  dynamic agents whose trajectories are a priori unknown. Let $\mathcal{D}$ be an unknown distribution over agent trajectories and let 
\begin{align*}
    (Y_0,Y_1,\hdots)\sim \mathcal{D}
\end{align*} 
describe a random trajectory where the joint agent state $Y_t:=(Y_{t,1},\hdots,Y_{t,N})$ at time $t$ is drawn from $\mathbb{R}^{N  n}$, i.e., $Y_{t,j}$ is the state of agent $j$ at time $t$.\footnote{For simplicity, we assume that the state of each dynamic agent is $n$-dimensional.  This assumption can easily be generalized.} We assume at time $t$ to have knowledge of  $(Y_0,\hdots,Y_t)$. Modeling dynamic agents by a distribution $\mathcal{D}$ provides great flexibility, e.g., the pedestrians in Fig. \ref{fig:intro_figure} can be described by distributions $\mathcal{D}_1$, $\mathcal{D}_2$, and $\mathcal{D}_3$ with joint distribution $\mathcal{D}$, and $\mathcal{D}$ can generally describe the motion of Markov decision processes. We make no assumptions on the form of the distribution  $\mathcal{D}$, but assume that $\mathcal{D}$ is independent of the system in~\eqref{eq:system} as formalized next.
 \begin{assumption}\label{rem:1}
For any time $t\ge 0$, the control inputs $(u_0,\hdots,u_{t-1})$ and the resulting trajectory $(x_0,\hdots, x_t)$, following \eqref{eq:system}, do not change the distribution of $(Y_0,Y_1,\hdots)\sim\mathcal{D}$. 
\end{assumption}

Assumption \ref{rem:1} approximately holds in many applications, e.g., a self-driving car taking conservative control actions that result in socially acceptable trajectories which do not change the behavior of pedestrians. We later comment on ways to deal with distribution shifts in practice, and reserve a thorough treatment of this issue for future papers. We further assume availability of training and calibration data drawn from $\mathcal{D}$. 
 \begin{assumption}\label{ass1}
 	We have  a dataset $D:=\{Y^{(1)},\hdots,Y^{(K)}\}$ in which each of the $K$ trajectories $Y^{(i)}:=(Y_0^{(i)},Y_1^{(i)},\hdots)$ is independently drawn from $\mathcal{D}$, i.e., $Y^{(i)}\sim\mathcal{D}$.
 \end{assumption}

 Assumption \ref{ass1} is not restrictive in practice, e.g., pedestrian data is available in autonomous driving.  Let us now define the problem that we aim to solve in this paper.

\begin{problem}\label{prob1}	Given the system in \eqref{eq:system}, the random trajectory $(Y_0,Y_1\hdots)\sim\mathcal{D}$, a mission time $T$,  and a failure probability $\delta\in(0,1)$, design the control inputs $u_t$ such that the Lipschitz continuous constraint function $c:\mathbb{R}^n\times\mathbb{R}^{n N}\to \mathbb{R}$ is satisfied\footnote{We assume that $c$ is initially satisfied, i.e., that $c(x_0,Y_0,0)\ge 0$.} with a probability of at least $1-\delta$, i.e., that
\vspace{-0.1cm}
\begin{align*}
    P\big(c(x_t,Y_t)\ge 0, \forall t\in\{0,\hdots,T\}\big)\ge 1-\delta.
\end{align*}
\end{problem}

We note that the function $c$ can encode collision avoidance, but also  objectives such as tracking another agent.  In our solution to Problem \ref{prob1}, we additionally achieve cost optimality in terms of a cost function $J$ (details below).

\subsection{Trajectory Predictors for Dynamic Environments}
\label{sec:RNN}

Our goal is to predict future agent states $(Y_{t+1},\hdots,Y_{T})$  from observations $(Y_0,\hdots,Y_t)$. Our proposed planning algorithm is compatible with any trajectory prediction algorithm.  Assume  that \textsc{Predict} is a measureable function that maps observations $(Y_0,\hdots,Y_t)$ to predictions $(\hat{Y}_{t+1|t},\hdots,\hat{Y}_{T|t})$ of $(Y_{t+1},\hdots,Y_{T})$. We now split the dataset $D$ into training and calibration datasets $D_\text{train}$ and $D_\text{cal}$, respectively, and assume that \textsc{Predict} is learned from $D_{\text{train}}$. 
 
A specific example of \textsc{Predict} are recurrent neural networks (RNNs) that have shown good performance \cite{rudenko2020human}.  For $\tau\le t$, the recurrent structure of an RNN is given as 
\begin{align*}
     h_\tau^1&:=\mathcal{H}(Y_\tau,h_{\tau-1}^1), \;\;\;\;\;\;\;\;\;\;\;\; \\
     h_\tau^i&:=\mathcal{H}(Y_\tau,h_{\tau-1}^{i},h_\tau^{i-1}), \;\;\;\; \forall i\in\{2,\hdots,d\}\\
     \hat{Y}_{\tau+1|\tau}&:=\mathcal{Y}(h_\tau^d),
 \end{align*}
 where $Y_\tau$ is the input that is sequentially applied and  $\mathcal{H}$ is a function that can parameterize different types of RNNs, e.g., LSTMs  \cite{hochreiter1997long}. Furthermore, $d$ is its depth and $h_\tau^1,\hdots,h_\tau^d$ are the hidden states. The output $\hat{Y}_{t+1|t}$  provides an estimate of $Y_{t+1}$, see e.g., \cite{graves2013generating,salzmann2020trajectron} where the function $\mathcal{Y}$ parameterizes a predictive conditional distribution. To obtain the remaining predictions $\hat{Y}_{t+2|t},\hdots,\hat{Y}_{T|t}$ of $Y_{t+2},\hdots,Y_T$, we use the RNN in a recursive way by sequentially applying $\hat{Y}_{t+1|t},\hdots,\hat{Y}_{T-1|t}$ instead of the unknown  ${Y}_{t+1},\hdots,{Y}_{T-1}$.

\section{Conformal Prediction Regions for Trajectory Predictors}

The challenges in solving Problem \ref{prob1} are twofold. First, the unknown distribution $\mathcal{D}$ over trajectories $(Y_0,Y_1,\hdots)$ can be complex and may not follow standard assumptions, e.g., being Gaussian. Second, the function \textsc{Predict} can be highly nonlinear and predictions $\hat{Y}_{\tau|t}$ may not be accurate. To be able to deal with these  challenges, we use conformal prediction to obtain \emph{prediction regions} for  $\hat{Y}_{\tau|t}$.  

\subsection{Introduction to Conformal Prediction}
\label{sec:intro_conf}
Conformal prediction was introduced  in \cite{vovk2005algorithmic,shafer2008tutorial} to obtain valid prediction regions for complex predictive models, i.e., neural networks, without making assumptions on the underlying distribution or the predictive model \cite{angelopoulos2021gentle,zeni2020conformal,lei2018distribution,tibshirani2019conformal,cauchois2020robust}.  We first provide a brief introduction to conformal prediction. 

Let $R^{(0)},\hdots,R^{(k)}$ be $k+1$ independent and identically distributed random variables.\footnote{In general only exchangeability is needed, which means that the joint distribution of $R^{(0)},\hdots,R^{(k)}$ is the same as the joint  distribution of $R^{(\sigma(0))},\hdots,R^{(\sigma(k))}$ for any permutation $\sigma$ on $\{0,\hdots k\}$. Exchangeability is a strictly weaker requirement than independence and identical distribution.} The variable $R^{(i)}$ is usually referred to as the \emph{nonconformity score}. In supervised learning, it may be defined as $R^{(i)}:=\|Z^{(i)}-\mu(X^{(i)})\|$ where the predictor $\mu$ attempts to predict the output  $Z^{(i)}$ based on the input $X^{(i)}$.  A large nonconformity score indicates a poor predictive model. Our goal is to obtain a prediction region for $R^{(0)}$ based on $R^{(1)},\hdots,R^{(k)}$, i.e., the random variable $R^{(0)}$ should be contained within the prediction region  with high probability. Formally, given a failure probability $\delta\in (0,1)$, we want to construct a valid prediction region $C$ so that\footnote{More formally, we would have to write $C(R^{(1)},\hdots,R^{(k)})$ as the prediction region $C$ is a function of $R^{(1)},\hdots,R^{(k)}$. For this reason, the probability measure $P$ is defined over the product measure of $R^{(0)},\hdots,R^{(k)}$.}
\begin{align*}
    P(R^{(0)}\le C)\ge 1-\delta.
\end{align*}

By a surprisingly simple quantile argument, see \cite[Lemma 1]{tibshirani2019conformal}, one can obtain $C$ to be the $(1-\delta)$th quantile of the empirical distribution of the values $R^{(1)},\hdots,R^{(k)}$ and $\infty$. By assuming that $R^{(1)},\hdots,R^{(k)}$ are sorted in non-decreasing order, and by adding $R^{(k+1)}:=\infty$, we can equivalently obtain $C:=R^{(p)}$ where $p:=\lceil (k+1)(1-\delta)\rceil$ \textcolor{black}{ with $\lceil \cdot\rceil$ being the ceiling function}, i.e., $C$ is the $p$th smallest nonconformity score. In the next section, we discuss how to obtain conformal prediction regions for trajectory predictions. We are inspired by \cite{stankeviciute2021conformal}  where prediction regions for RNNs were derived, but present results that are more taylored for our problem setup.

\subsection{Prediction Regions using Conformal Prediction}

Given observations $(Y_0,\hdots,Y_t)$ at time $t$, recall that we can obtain predictions $\hat{Y}_{\tau|t}$ of $Y_\tau$ for all future times $\tau\in\{t+1,\hdots,T\}$ via the \textsc{Predict} function. For a failure probability of $\bar{\delta}\in(0,1)$, our first goal is now to construct prediction regions defined by a value $C_{\tau|t}$ so that
\begin{align*}
    P\big(\|Y_\tau-\hat{Y}_{\tau|t}\|\le C_{\tau|t})\ge 1-\bar{\delta}.
\end{align*}
Following Section \ref{sec:intro_conf}, we define the nonconformity score $R_{\tau|t}:=\|Y_\tau-\hat{Y}_{\tau|t}\|$ so that a small (large) nonconformity score  indicates that the predictions $\hat{Y}_{\tau|t}$ are accurate (not accurate). Indeed, let us compute the nonconformity score for each trajectory $Y^{(i)}$ of the calibration dataset $D_\text{cal}$  as
\begin{align*}
    R_{\tau|t}^{(i)}:=\|Y_\tau^{(i)}-\hat{Y}_{\tau|t}^{(i)}\|
\end{align*}
where $\hat{Y}_{\tau|t}^{(i)}$ is the prediction obtained from  $(Y^{(i)}_0,\hdots,Y^{(i)}_t)$.

We can now  obtain prediction regions by a direct application of \cite[Lemma 1]{tibshirani2019conformal}. Assume hence that the values of $R_{\tau|t}^{(i)}$ are sorted in non-decreasing order, and let us add  $R_{\tau|t}^{(|D_\text{cal}|+1)}:=\infty$ as the $(|D_\text{cal}|+1)$th value.

\begin{lemma}[Following \cite{vovk2005algorithmic}]\label{lem:1}
Given the random trajectory $(Y_0,Y_1,\hdots)\sim\mathcal{D}$, the calibration dataset $D_\text{cal}$, and  predictions $\hat{Y}_{\tau|t}$ obtained from  observations $(Y_0,\hdots,Y_t)$. Let $\bar{\delta}\in(0,1)$ be a failure probability, then for $\tau>t$ we have
\begin{align*}
    P\big(\|Y_\tau-\hat{Y}_{\tau|t}\|\le C_{\tau|t})\ge 1-\bar{\delta}
\end{align*}
where the prediction regions $C_{\tau|t}$ are defined by
\begin{align}\label{eq:lem_1}
    C_{\tau|t}:=R_{\tau|t}^{(p)}\;\;\text{with}\;\;  p:=\big\lceil (|D_\text{cal}|+1)(1-\bar{\delta})\big\rceil.
\end{align}
\begin{proof}
By construction of the nonconformity scores $R_{\tau|t}^{(i)}$, they are exchangeable and we can directly apply \cite[Lemma 1]{tibshirani2019conformal} so that $P\big(\|Y_\tau-\hat{Y}_{\tau|t}\|\le C_{\tau|t})\ge 1-\bar{\delta}$.
\end{proof}
\end{lemma}

We can now construct prediction regions over multiple future predictions by an application of Boole's inequality. 
\begin{theorem}\label{thm:1}
Given the random trajectory $(Y_0,Y_1,\hdots)\sim\mathcal{D}$, the calibration dataset $D_\text{cal}$, and predictions $\hat{Y}_{\tau|t}$ obtained from observations $(Y_0,\hdots,Y_t)$. Let $\bar{\delta}:=\delta/T$ where $\delta\in(0,1)$ is a failure probability, then the following two statements hold where $C_{\tau|t}$ is constructed as in \eqref{eq:lem_1}:
\begin{align}
    &P\big(\|Y_\tau-\hat{Y}_{\tau|0}\|\le C_{\tau|0}, \;\forall\tau\in\{1,\hdots,T\})\ge 1-\delta,\label{eq:thm1_1}\\
    &P\big(\|Y_{t+1}-\hat{Y}_{t+1|t}\|\le C_{t+1|t}, \;\forall t\in\{0,\hdots,T-1\})\ge 1-\delta.\label{eq:thm1_2}
\end{align}
\begin{proof}
Let us first show that the  statement in equation \eqref{eq:thm1_1} holds. According to Lemma \ref{lem:1}, it holds that $P\big(\|Y_\tau-\hat{Y}_{\tau|0}\|\le C_{\tau|0})\ge 1-\bar{\delta}$ for each $\tau\in\{1,\hdots,T\}$ individually. \vspace{0.2cm} We consequently know that $P\big(\|Y_\tau-\hat{Y}_{\tau|0}\|> C_{\tau|0})\le \bar{\delta}$. Applying Boole's inequality gives us
\begin{align*}
    P\big(\exists \tau>0 \text{ s.t. }\|Y_\tau-\hat{Y}_{\tau|0}\|> C_{\tau|0})\le\sum_{i=t}^T\bar{\delta}= \sum_{i=t}^T\frac{\delta}{T}=\delta
\end{align*}
so that we can finally conclude that
\begin{align*}
    P\big(\|Y_\tau-\hat{Y}_{\tau|0}\|\le C_{\tau|0},\; \forall \tau\in\{1,\hdots,T\})\ge 1- \delta
\end{align*}
which proves \eqref{eq:thm1_1}. Statement \eqref{eq:thm1_2} follows analogously.
\end{proof}
\end{theorem}

\textcolor{black}{Equation \eqref{eq:thm1_1} guarantees that all $\tau$-step ahead prediction regions at time zero are valid, while equation \eqref{eq:thm1_2} guarantees that all one-step ahead prediction regions are valid. We note that we set $\bar{\delta}:=\delta/T$ to compute $C_{\tau|t}$ according to equation \eqref{eq:lem_1} to obtain prediction regions over multiple time steps. As a consequence, the value of $C_{\tau|t}$ increases with increasing $T$ or decreasing $\delta$ as smaller $\bar{\delta}$ result in larger quantiles $p$.} 

\begin{remark}
The result in equation \eqref{eq:thm1_1} was similarly shown in \cite{stankeviciute2021conformal}, but without permitting recursive RNNs, i.e., RNNs where predictions $\hat{Y}_{\tau|t}$ are recursively used to predict $\hat{Y}_{\tau+1|t}$. We  permit any measurable predictor \textsc{Predict}.  Importantly, we also show that equation \eqref{eq:thm1_2} holds, which will be important  for proving correctness guarantees of our proposed MPC.
\end{remark}
\begin{remark}\label{C_max}
For the applications that we plan to address, it can be useful to instead consider a nonconformity score
\begin{align*}
    R_{\tau|t}:=\max_{j\in\{1,\hdots,N\}}\|Y_{\tau,j}-\hat{Y}_{\tau|t,j}\|
\end{align*}
where we recall that $Y_{\tau,j}$ is the state of agent $j$ at time $\tau$, and where $\hat{Y}_{\tau|t,j}$ is the corresponding prediction for agent $j$. We can  obtain prediction regions $P\big(R_{\tau|t}\le C_{\tau|t})\ge 1-\bar{\delta}$ in the same way as described above. Using this nonconformity score, we get prediction regions individually for each agent.
\end{remark}

\section{Model Predictive Control with Conformal Prediction Regions}

 We next propose an MPC that uses the predictions $\hat{Y}_{\tau|t}$ and the  prediction regions defined by $C_{\tau|t}$ to solve Problem~\ref{prob1}. Let us first present the optimization problem that will be iteratively solved within the MPC. Therefore, denote the Lipschitz constant of $c$ by $L$. For instance, the collision avoidance constraint $c(x,y):=\|x-y\|-0.5$ has Lipschitz constant one, i.e., $|c(x,y')-c(x,y'')|\le \|y'-y''\|$. At time $t$, we solve the following optimization problem to obtain an open-loop control sequence $u_t,\hdots,u_{T-1}$:
\begin{subequations}\label{eq:open_loop}
\begin{align}
    &\min_{(u_t,\hdots,u_{T-1})} J(x,u) &\\
     \text{s.t.}\;\;& x_{\tau+1}=f(x_\tau,u_\tau), &\tau\in\{t,\hdots,T-1\}\\
     & c(x_\tau,\hat{Y}_{\tau|t})\ge LC_{\tau|t},&\tau\in\{t+1,\hdots,t+H\}\label{eq:constC_2}\\
     & u_\tau \in \mathcal{U},x_{\tau+1} \in \mathcal{X},&\tau\in\{t,\hdots,T-1\} 
\end{align}
\end{subequations}
where $H$ is a prediction horizon and $J$ is a cost function over states $x:=(x_1,\hdots,x_T)$ and control inputs $u:=(u_0,\hdots,u_{T-1})$. The optimization problem \eqref{eq:open_loop} is convex if the functions $J$ and $f$ are convex, the function $c$ is convex in its first argument, and the sets $\mathcal{U}$ and $\mathcal{X}$ are convex. 

If one can solve the optimization problem \eqref{eq:open_loop} at time $t=0$, one obtains a control sequence $u$ that solves Problem \ref{prob1}. 
\begin{theorem}[Open-loop control]\label{thm:2}
Let the system  \eqref{eq:system} be given, and let the conditions from Theorem \ref{thm:1} and Assumption \ref{rem:1} hold.  If the optimization problem \eqref{eq:open_loop} is feasible at time $t=0$ with prediction horizon $H=T$, then the open-loop control sequence $u$ from \eqref{eq:open_loop} is s.t.
\begin{align*}
    P\big(c(x_\tau,Y_\tau)\ge 0, \forall \tau\in\{1,\hdots,T\}\big)\ge 1-\delta.
\end{align*}
\begin{proof}
Due to constraint \eqref{eq:constC_2}, it holds that
\begin{align*}
    0&\le c(x_\tau,\hat{Y}_{\tau|0})- LC_{\tau|0}\\
    &\le c(x_\tau,Y_\tau)+L\|Y_\tau-\hat{Y}_{\tau|0}\|- LC_{\tau|0}
\end{align*}
where the latter inequality follows by Lipschitz continuity. Now, by equation \eqref{eq:thm1_1} in Theorem \ref{thm:1}, it directly follows that $P\big(c(x_\tau,Y_\tau)\ge 0, \forall \tau\in\{1,\hdots,T\}\big)\ge 1-\delta$.
\end{proof}
\end{theorem}

While the open-loop controller in Theorem \ref{thm:2} provides a solution to Problem \ref{prob1}, the controller will admittedly be conservative, or the optimization problem in \eqref{eq:open_loop} may even be infeasible. The main reason for this is that the prediction regions defined by $C_{\tau|0}$ will be large for large $\tau$ as the predictions $\hat{Y}_{\tau|0}$ will lose accuracy. Another drawback of an open-loop controller is the missing robustness due to the lack of feedback. We propose a receding horizon control strategy in Algorithm \ref{alg:overview} that reduces conservatism and is robust.

\begin{algorithm}
    \centering
    \begin{algorithmic}[1]
        \Statex \textbf{Input: } Failure probability $\delta$, calibration dataset $D_\text{cal}$ obtained from $\mathcal{D}$, prediction and task horizons $H$ and~$T$
        \Statex \textbf{Output: } Feedback controller $u_0(x_0),\hdots,u_{T-1}(x_{T-1})$ 
        \State $\bar{\delta} \gets \delta/T$ 
        \State $p \gets \big\lceil (|D_\text{cal}|+1)(1-\bar{\delta})\big\rceil$
        \FOR{$t$ from $0$ to $T-1$} \quad\# offline computation
            \FOR{$\tau$ from $t+1$ to $t+H$} \quad\# conformal prediction
            \State Obtain predictions $\hat{Y}_{\tau|t}^{(i)}$ for each $Y^{(i)}\in D_\text{cal}$
            \State $R_{\tau|t}^{(i)} \gets \|Y_\tau^{(i)}-\hat{Y}_{\tau|t}^{(i)}\|$ for each $Y^{(i)}\in D_\text{cal}$
            \State $R_{\tau|t}^{(|D_\text{cal}|+1)} \gets \infty$
            \State Sort  $R_{\tau|t}^{(i)}$ in non-decreasing order
            \State $C_{\tau|t} \gets R_{\tau|t}^{(p)}$
            \ENDFOR
        \ENDFOR
        \FOR{$t$ from $0$ to $T-1$} \quad\# real-time planning loop
            \State Sense $x_t$ and  $Y_t$ 
            \State Obtain predictions $\hat{Y}_{\tau|t}$ for $\tau\in\{t+1,\hdots,t+H\}$
            \State Calculate controls $u_t,...,u_{T-1}$ as the solution of \eqref{eq:open_loop}
            \State Apply $u_t$ to \eqref{eq:system}
        \ENDFOR
    \end{algorithmic}
    \caption{MPC with Conformal Prediction Regions}
    \label{alg:overview}
\end{algorithm}

In lines 1 and 2 of Algorithm \ref{alg:overview}, we set the variables $\bar{\delta}$ and $p$ according to Lemma \ref{lem:1} and Theorem \ref{thm:1}. Lines 3-9 present the computation of the conformal prediction regions by: 1) calculating the predictions $\hat{Y}_{\tau|t}^{(i)}$ on the calibration data $D_\text{cal}$, 2) calculating the nonconformity scores $R_{\tau|t}^{(i)}$, and 3) obtaining $C_{\tau|t}:=R_{\tau|t}^{(p)}$ according to Theorem \ref{thm:1}. Lines 10-14 are the real-time planning loop in which we observe our states $x_t$ and $Y_t$ (line 11), obtain new predictions $\hat{Y}_{\tau|t}$ based on $Y_t$ (line 12), and solve the optimization problem in \eqref{eq:open_loop} of which we apply only $u_t$ (lines 13-14). The MPC presented in Algorithm \ref{alg:overview} enjoys the following guarantees. 

\begin{theorem}[Closed-loop control]\label{thm:3}
Let the system  \eqref{eq:system} be given, and let the conditions from Theorem \ref{thm:1} and Assumption \ref{rem:1} hold.  If the optimization problem \eqref{eq:open_loop} is feasible at each time $t\in \{0,\hdots,T-1\}$, then the closed-loop control $u_0(x_0),\hdots,u_{T-1}(x_{T-1})$ obtained from Algorithm \ref{alg:overview} is s.t.
\begin{align*}
    P\big(c(x_t,Y_t)\ge 0, \forall t\in\{1,\hdots,T\}\big)\ge 1-\delta.
\end{align*}

\begin{proof}
By assumption, the optimization problem \eqref{eq:open_loop} is feasible at each time $t\in \{0,\hdots,T-1\}$. Due to constraint \eqref{eq:constC_2} and Lipschitz continuity of $c$, it hence holds that
\begin{align*}
    0&\le c(x_{t+1},\hat{Y}_{t+1|t})- LC_{t+1|t}\\
    &\le c(x_{t+1},Y_{t+1})+L\|Y_{t+1}-\hat{Y}_{{t+1}|t}\|- LC_{{t+1}|t}
\end{align*}
at each time $t\in \{0,\hdots,T-1\}$.
By equation \eqref{eq:thm1_2} in Theorem \ref{thm:1}, it now follows that $P\big(c(x_{t+1},Y_{t+1})\ge 0, \forall t\in\{0,\hdots,T-1\}\big)\ge 1-\delta$ which proves the main result.
\end{proof}
\end{theorem}

We conclude  with final remarks on parameter choices and recursive feasibility of the MPC and distribution shifts in $\mathcal{D}$.
\begin{remark}
\textcolor{black}{A smaller failure probability $\delta$  leads to larger prediction regions $C_{\tau|t}$, as remarked before, and hence to less optimal paths w.r.t. the cost function $J$. Larger prediction horizons $H$ may also lead to less optimal paths as future predictions lose accuracy. For too small $H$, however, one may experience recursive feasibility issues of the optimization problem \eqref{eq:open_loop}. While we assume recursive feasibility in Theorem \ref{thm:3}, we note that this is a reasonable assumption when $\|Y_{t+H}-Y_{t+H+1}\|$ is not changing too much. Note that the time-varying nature of prediction regions is typically no problem for recursive feasibility as prediction regions shrink over time, e.g., $C_{\tau|t}$ will typically be smaller than $C_{\tau+1|t}$.}
\end{remark}

\begin{remark}
\textcolor{black}{We assume that test and calibration trajectories follow the same distribution $\mathcal{D}$ per Assumptions \ref{rem:1} and \ref{ass1}.  We hence ignore that the trajectory $x$ may change $\mathcal{D}$ during test time, e.g., when a robot is too close to a pedestrian.} While we do not address distribution shifts in full generality, e.g., when  $\mathcal{D}(x)$ depends explicitly on $x$, we can use robust conformal prediction \cite{cauchois2020robust} to obtain valid prediction regions for all distributions that are ``close'' to $\mathcal{D}$ (in terms of the f-divergence), and integrate these in an MPC.
\end{remark}

\section{Case Studies}
\label{sec:simulations}

In the first case study, we consider navigating a mobile robot around \textcolor{black}{pedestrians whose trajectories are generated in TrajNet++ \cite{kothari2021human} using the ORCA simulator \cite{van2008reciprocal}}. In the second case study, we control a self-driving car at an intersection filled with  pedestrians in CARLA \cite{Dosovitskiy17}, see Fig.~\ref{fig:intro_figure}.

For trajectory prediction, we use the social LSTM from \cite{alahi2016social}. Compared to vanilla LSTMs as introduced in Section \ref{sec:RNN}, the social LSTM uses one LSTM for each  agent while sharing LSTM weights via a pooling layer to model social interactions. These pooled LSTM weights are then used as inputs of the individual LSTMs,  see \cite{alahi2016social} for details. For both case studies, we trained a social LSTM with a depth of $d:=128$ for the individual agent LSTM.

In both case studies, we consider a bicycle model \cite{pepy2006path}
\begin{align*}
    \begin{bmatrix}
    x_{t+1}\\
    y_{t+1}\\
    \theta_{t+1}\\
    v_{t+1}
    \end{bmatrix}=\begin{bmatrix}
    x_t + \Delta v_t \cos(\theta_t)\\
    x_t + \Delta v_t \sin(\theta_t)\\
    \theta_t + \Delta \frac{v_t}{l} \tan(\phi_t)\\
    v_t + \Delta a_t
    \end{bmatrix}
\end{align*}
where $p_\tau:=(x_t,y_t)$ is the two-dimensional position, $\theta_t$ is the vehicles's orientation, $v_t$ is the velocity, $l:=1$ is the vehicles's length, and $\Delta$ is the sampling time. The control inputs are the steering angle $\phi_t$ and the acceleration $a_t$.

The  objective is to reach a goal region, while avoiding the pedestrians by means of the constraint function 
\begin{align*}
    c(p_\tau,Y_\tau):=\min_{j\in \{1,\hdots,N\}}\|p_\tau-Y_{\tau,j}\|-\epsilon
\end{align*}
 where $\epsilon$ is a user-defined safety distance. In the first case study, we encode reaching the goal region as a constraint $\|p_T-p_\text{goal}\|\le 0.25$, while we minimize $\|p_T-p_\text{goal}\|$ in the cost function $J$ in the second case study. \textcolor{black}{To solve the optimization problem in \eqref{eq:open_loop}, we use CasADi \cite{andersson2019casadi} with the Ipopt nonlinear programming  solver.} Animations for both case studies can be found at \href{https://tinyurl.com/ecz2a9c4}{https://tinyurl.com/ecz2a9c4}.

\begin{figure}
        \vspace{0.12cm}
		\centering
		\hspace{-0.5cm}\includegraphics[scale=0.06]{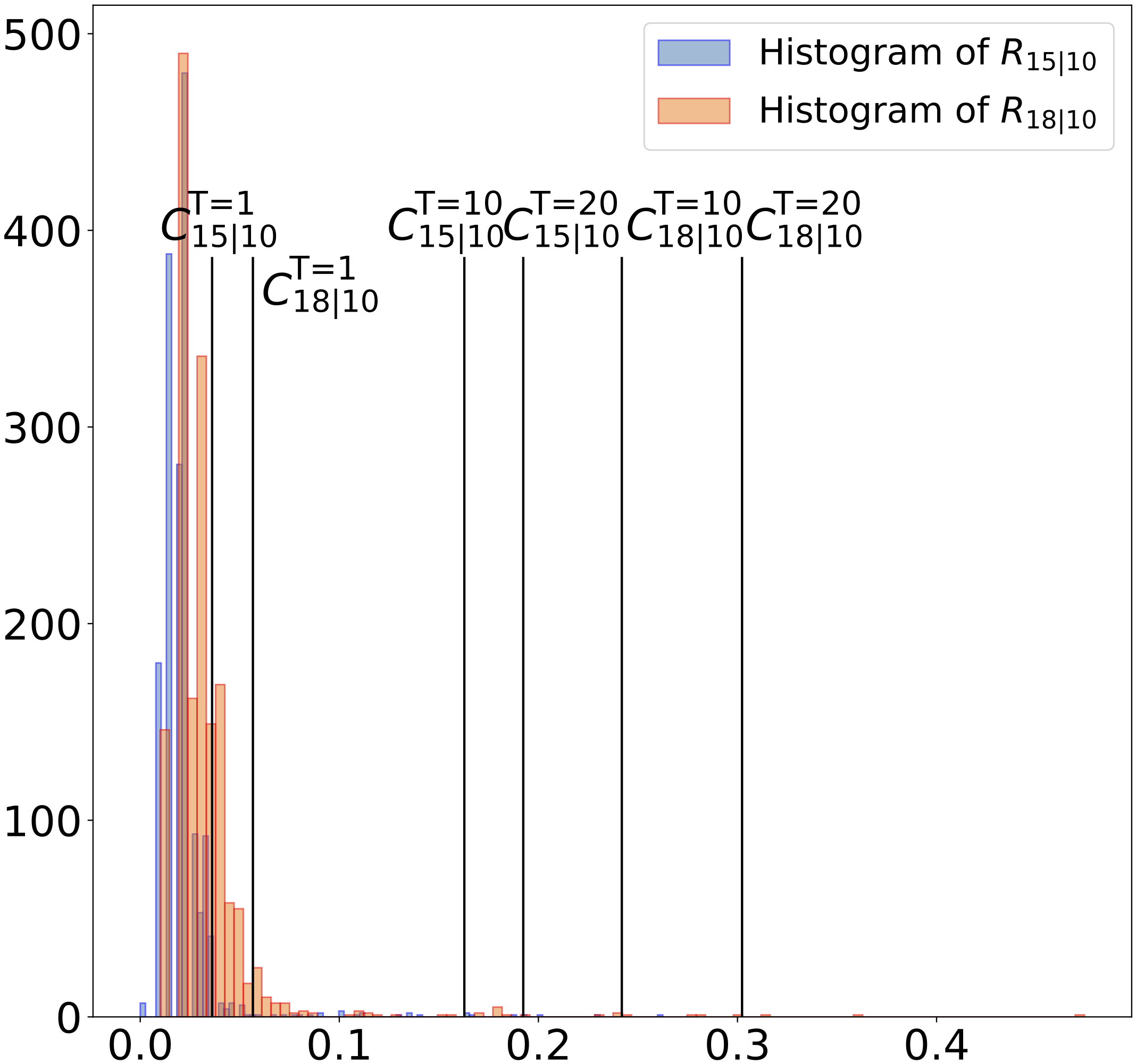}\hspace{0.125cm}
		\includegraphics[scale=0.06]{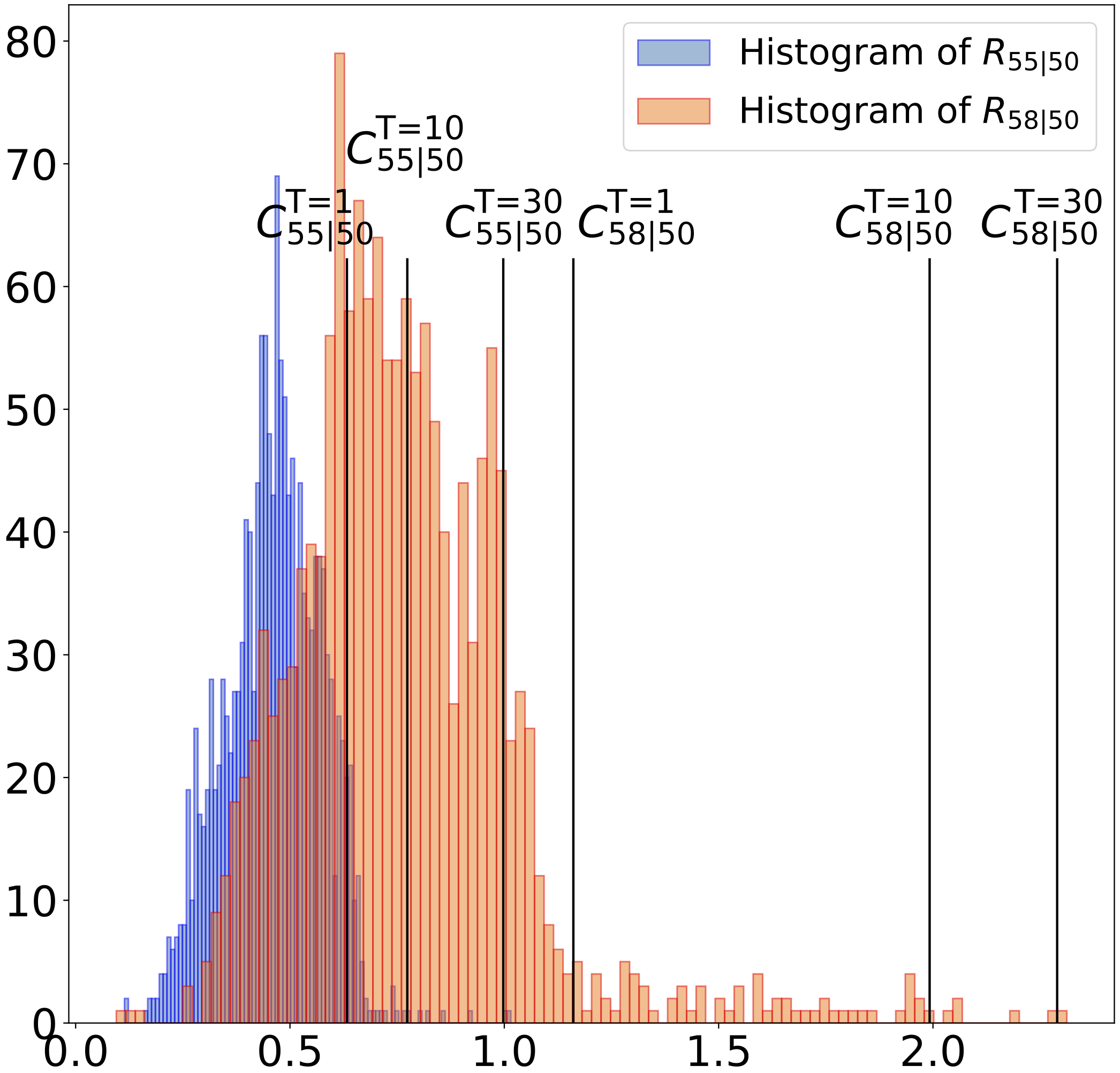}
		\caption{\textcolor{black}{Left: nonconformity scores $R_{15|10}^{(i)}$ and $R_{18|10}^{(i)}$ on $D_\text{cal}$ along with $C_{15|10}$ and $C_{18|10}$ for different $T$ for Scenario 1. Right: nonconformity scores $R_{55|50}^{(i)}$ and $R_{58|50}^{(i)}$ on $D_\text{cal}$ along with $C_{55|50}$ and $C_{58|50}$ for different $T$ for Scenario 2.}}
		\label{fig:scen1_hist}
		\vspace{-0.5cm}
\end{figure}


\begin{figure*}
\centering
\vspace{-0.15cm}
\includegraphics[scale=0.375]{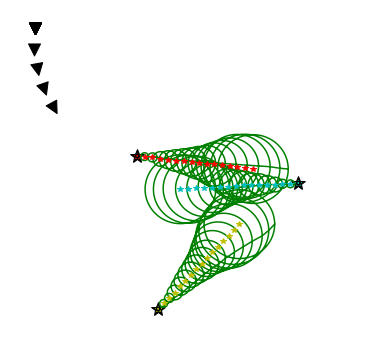}
\includegraphics[scale=0.375]{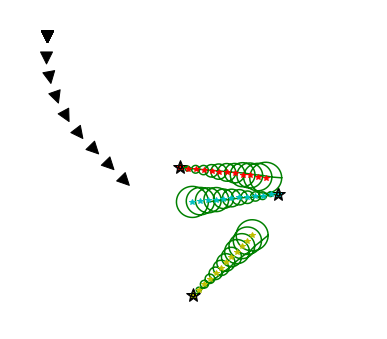}
\includegraphics[scale=0.375]{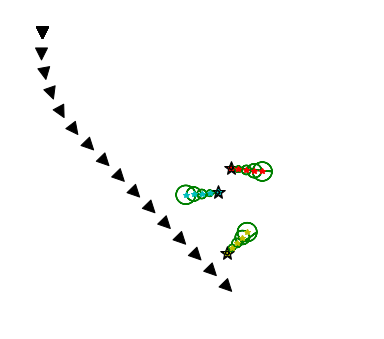}\\\vspace{-0.35cm}
\includegraphics[scale=0.375]{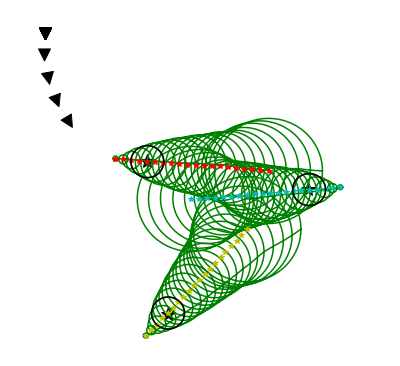}\vspace{-0.5cm}
\includegraphics[scale=0.375]{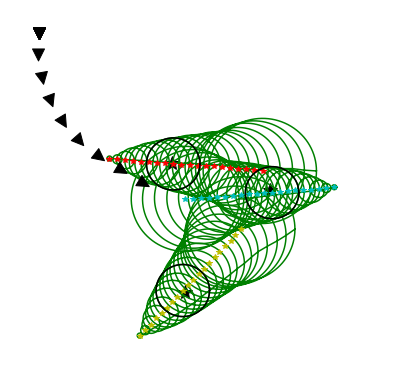}
\includegraphics[scale=0.375]{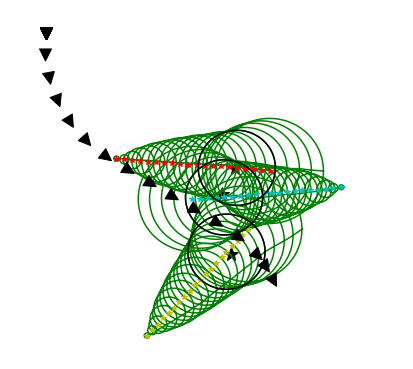}
\caption{Scenario 1. The top three plots show the robot trajectory (black triangles) produced by the proposed MPC  at times $4$, $8$, and $15$. There are three pedestrians, with their actual trajectories indicated by black stars, that the robot safely avoids. The LSTM predictions are indicated by red stars with the corresponding prediction regions shown in green. The prediction regions are updated at each time making the MPC less conservative than the open-loop controller, which is presented in the bottom three plots. For the open-loop controller, the prediction regions are fixed from time zero and given by $C_{\tau|0}$. }
\label{fig:scen1}
\vspace{-0.5cm}
\end{figure*}

\textbf{Scenario 1 (ORCA). } We collected $4500$ synthetic trajectories in a scene consisting of three pedestrians, and created training, calibration, and test datasets with sizes $|D_\text{train}|=2000$,  $|D_\text{cal}|=2000$, and $|D_\text{test}|:=500$, respectively. The sampling time is $\Delta:=1/8$, and Fig. \ref{fig:scen1_hist} (left) shows histograms of the nonconformity scores $R_{\tau|t}^{(i)}$ evaluated on $D_\text{cal}$ for $\tau\in\{15,18\}$ and $t:=10$ (with $R$ defined as in Remark \ref{C_max}). Based on these nonconformity scores, we can calculate $C_{\tau|t}$ according to Theorem \ref{thm:1} by using $\delta:=0.05$ and $T:=20$. In Fig. \ref{fig:scen1_hist} (left), we additionally indicate $C_{\tau|t}$ for different values of $T$, and we can observe that larger $T$ naturally result in larger prediction regions. Next, we empirically evaluate the correctness of the prediction regions by checking whether or not equation \eqref{eq:thm1_1} holds on $D_\text{test}$. Indeed, we find that $498$ of the $500$ test trajectories are such that $\|Y_\tau^{(i)}-\hat{Y}_{\tau|0}^{(i)}\|\le C_{\tau|0}$ which empirically confirms equation \eqref{eq:thm1_1} in Theorem \ref{thm:1}. For one of these test trajectories, Fig. \ref{fig:scen1} shows the prediction regions defined by $\hat{Y}_{\tau|t}$ and $C_{\tau|t}$ for $\tau>t$ where $t=0$ (bottom plots) $t=4$ (top left plot), $t=8$ (top middle plot), $t=15$ (top right plot). We note that the LSTM at time $t=0$ for simplicity uses information $Y_{-20},\hdots,Y_{-1}$.  We can observe that prediction regions become larger for larger $\tau$.

Fig. \ref{fig:scen1} shows the result of the proposed MPC (here with $H:=T$) in the top three plots for a trajectory $(Y_0,Y_1,\hdots)$ from $D_\text{test}$. For comparison, we present the results for the open-loop controller from Theorem \ref{thm:2}, i.e., only applying the control sequence obtained at time zero, in the bottom three plots of Fig. \ref{fig:scen1}. For illustration purposes, we have set $\epsilon:=0$. It is visible that both the \textcolor{black}{open and closed loop MPC controllers} are such that the pedestrians are avoided according to the constraint function $c$. This has to hold in at least $95$ percent of the cases by Theorems \ref{thm:2} and \ref{thm:3} since we have selected $\delta=0.05$. The cost function $J$ in the \textcolor{black}{open and closed loop MPC controllers} additionally minimizes $\sum_{\tau=1}^T v_\tau^2$ to avoid large velocities, which also acts as a proxy to minimize the  total distance traveled. The \textcolor{black}{closed loop} MPC is, as expected and motivated before, much less conservative than the open-loop controller and finds a more direct trajectory towards the goal region. To corroborate this observation, we calculated the averaged cost over $500$ test trajectories which are $414$ for the open-loop controller and $367.1$ for the \textcolor{black}{closed loop controller}.



\begin{figure}
		\centering
		\includegraphics[scale=0.12]{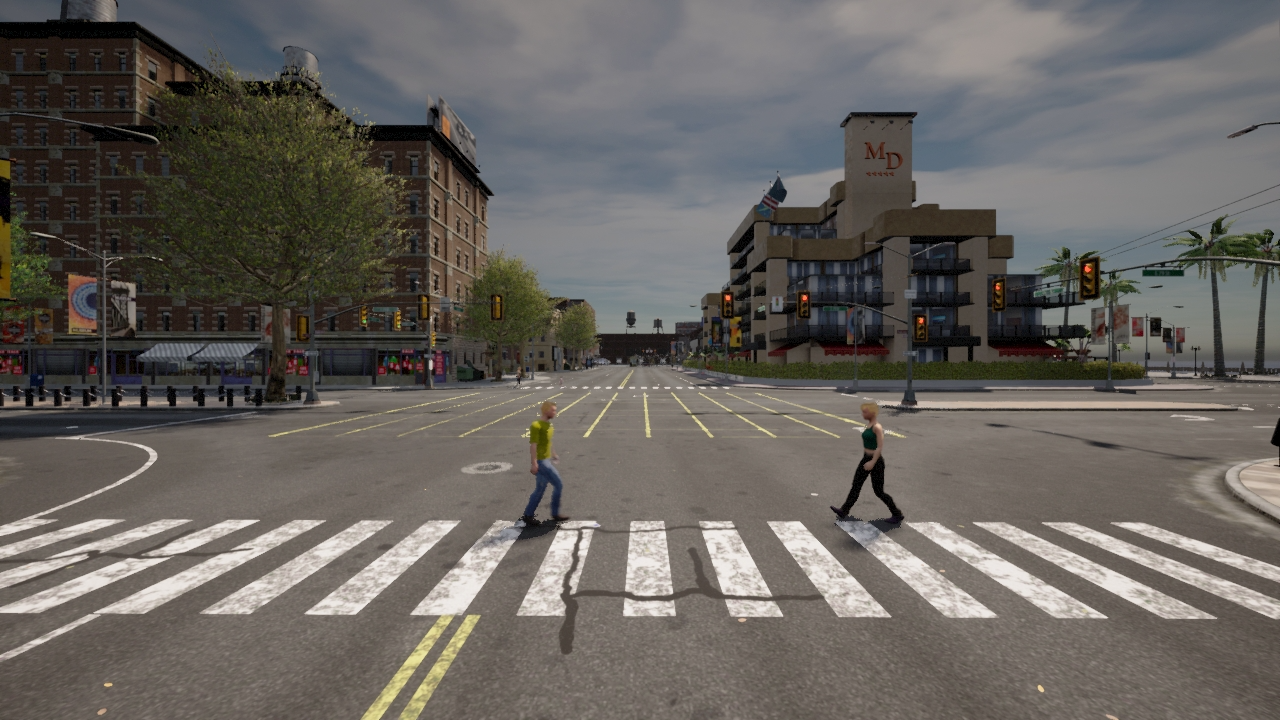}
		\caption{Intersection with pedestrians in CARLA \cite{Dosovitskiy17}.}
		\label{fig:intro_figured}
		\vspace{-0.2cm}
\end{figure}


\begin{figure*}
\centering
\vspace{0.04cm}
\includegraphics[scale=0.375]{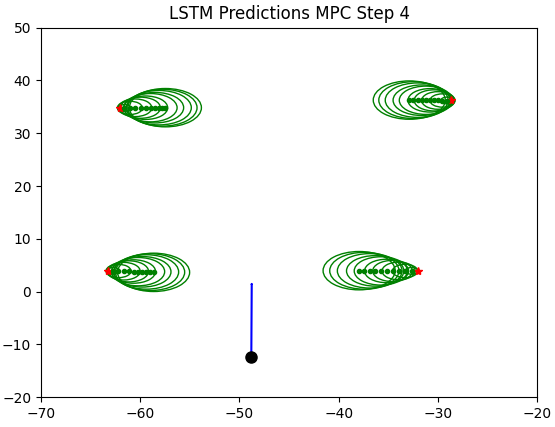}
\includegraphics[scale=0.375]{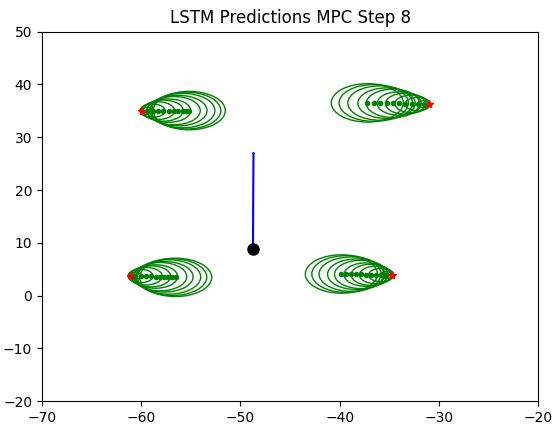}
\includegraphics[scale=0.375]{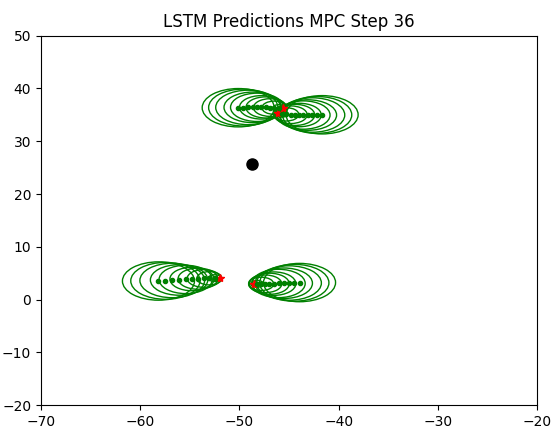}\\
\caption{Scenario 2. The car (black dot) navigating an intersection while avoiding four pedestrians (current position marked in red, predictions and prediction regions indicated by green dots and circles, respectively). Plots  at times $4$, $8$, and $36$.}
\label{fig:scen2}
\end{figure*}

\textbf{Scenario 2 (CARLA). } Within the autonomous driving simulator CARLA we have created a scene consisting of an intersection that is populated with four pedestrians, see Fig. \ref{fig:intro_figured}. These pedestrians have random initial and goal positions, and walk at random speeds to reach their goal. We collected $2600$ trajectories, and then created training, calibration, and test datasets with sizes $|D_\text{train}|=1000$, $|D_\text{cal}|=1500$, and $|D_\text{test}|=100$, respectively, and train an LSTM from $D_\text{train}$ to predict pedestrian trajectories.  We run  CARLA  at a frequency of $20$ Hz for $30$ seconds, but use the LSTM and solve the MPC at $2$ Hz. We show the histograms of the nonconformity scores $R_{\tau|t}^{(i)}$ evaluated on $D_\text{cal}$ for $\tau\in\{55,58\}$ and $t:=50$ which are hence the $2.5$ and $4$ seconds ahead prediction regions (with $R$ defined as in Remark \ref{C_max}) in Fig. \ref{fig:scen1_hist} (right). We again indicate $C_{\tau|t}$ for different values of $T$, and observe that the size of $C_{\tau|t}$ increases with larger $T$. As we run the car for $30$ seconds, the mission time $T$ is technically $60$. However, as we have a limited amount of calibration trajectories (data collection in CARLA is real-time), we only use $T:=30$ to calculate prediction regions $C_{\tau|t}$ from  $D_\text{cal}$, i.e., we use $\bar{\delta}:=\delta/30$ with $\delta=0.05$ in Theorem \ref{thm:1}. This choice is practically motivated and larger calibration datasets $D_\text{cal}$ will allow to select larger $T$. Nonetheless, the obtained prediction regions are valid for $T$ steps.

 We use the MPC again with the constraint function $c$ and the bicycle model discretized with $\Delta=1/2$, but we now set the prediction horizon to $H:=10$, i.e., to five seconds. Due to the large time horizon of $T$, we note that the open-loop controller with $H:=T$ is infeasible. We made two practically motivated modifications to the optimization problem in \eqref{eq:open_loop}. First, for the $\kappa$ step-ahead predictions we observed that the values of $C_{t+\kappa|t}$ are similar at different times $t$, and we have hence used the smallest value of $C_{t+\kappa|t}$ among all $t$ for the $\kappa$ step-ahead prediction region. Second, due to a model mismatch between the bicycle model and the CARLA model, we also included slack variables on the constraint in \eqref{eq:constC_2} to obtain recursive feasibility. 
 
The MPC result for a single trajectory from the test set $D_\text{test}$ is shown in Fig. \ref{fig:scen2}. We also ran the MPC for all $100$ test trajectories, and only in one case the safety constraint $c(x_\tau,Y_\tau)\ge 0$ was violated. This confirms Theorem \ref{thm:3} which states that in at most $5$ of the $100$ MPC runs the constraint is violated. We next checked the correctness of the one-step ahead prediction regions and found that in $99.985$ \% it holds that $\|Y_\tau-\hat{Y}_{\tau|t}\|\le C_{\tau|t}$ are satisfied which is  higher than the theoretically ensured value of $100(1-\delta/T)=99.83$ \%.

\section{Conclusion}
\label{sec:conclusion}
We presented a model predictive controller (MPC) that uses conformal prediction for probabilistic  safe planning in dynamic environments. The MPC uses complex trajectory predictors, such as (but not limited to) long short term memory networks, to predict future states of the dynamic environment and incorporates valid prediction regions using conformal prediction to quantify the uncertainty. To the best of our knowledge, these are the first results that provide valid safety guarantees for planning in dynamic environments without making assumptions on the predictor or the environment. To corroborate our results, we presented two numerical experiments of a mobile robot and a self-driving vehicle safely navigating around other agents.

For future work, we plan to do a comparative study using different trajectory predictors and to analyze their interplay with the MPC in more detail. We will also extend the presented MPC
to be able to handle an arbitrary number of agents in
the environment. We further plan to investigate potential conservatism in the prediction regions (induced by union bounding over events over several time steps in Theorem \ref{thm:1})  by using adaptive methods.

\section*{Acknowledgements}
This research was generously supported by NSF award CPS-2038873.

\bibliographystyle{IEEEtran}
\bibliography{literature}


\addtolength{\textheight}{-12cm}   

\end{document}